\documentclass[conference]{IEEEtran}
\usepackage{cite}
\usepackage{amsmath,amssymb,amsfonts}
\usepackage{algorithmic}
\usepackage{graphicx}
\usepackage{textcomp}
\usepackage{xcolor}

\usepackage{siunitx}
\usepackage{multirow}

\def\BibTeX{{\rm B\kern-.05em{\sc i\kern-.025em b}\kern-.08em
    T\kern-.1667em\lower.7ex\hbox{E}\kern-.125emX}}

\begin{document}

\title{Indoor Path Planning for Multiple Unmanned Aerial Vehicles via Curriculum Learning}

\author{\IEEEauthorblockN{Jongmin Park} 
\IEEEauthorblockA{\textit{School of Integrated Technology} \\
\textit{Yonsei University}\\
Incheon, Korea \\
jm97@yonsei.ac.kr} 
\and
\IEEEauthorblockN{Kwansik Park${}^{*}$} 
\IEEEauthorblockA{\textit{Korea Aerospace Research Institute} \\
Daejeon, Korea \\
kspark6469@kari.re.kr}
{\small${}^{*}$ Corresponding author}
}

\maketitle

\begin{abstract}
Multi-agent reinforcement learning was performed in this study for indoor path planning of two unmanned aerial vehicles (UAVs).
Each UAV performed the task of moving as fast as possible from a randomly paired initial position to a goal position in an environment with obstacles. To minimize training time and prevent the damage of UAVs, learning was performed by simulation. 
Considering the non-stationary characteristics of the multi-agent environment wherein the optimal behavior varies based on the actions of other agents, the action of the other UAV was also included in the state space of each UAV. Curriculum learning was performed in two stages to increase learning efficiency. A goal rate of 89.0\% was obtained compared with other learning strategies that obtained goal rates of 73.6\% and 79.9\%.
\end{abstract}


\section{Introduction}
The usage of unmanned aerial vehicles (UAVs) is widespread \cite{Alladi20:SecAuthUAV, Lee20191187, Singh20:On, Lee2018:Simulation}. Among them, the quadcopter \cite{Xuan-Mung20, Shin2017617, Talaeizadeh21, Kim20141431} is one of the most commonly used UAVs, and it has the advantage of being able to hover and rotate as required. There are various methods for estimating the location of a UAV, such as techniques utilizing global navigation satellite systems (GNSS) \cite{Causa21, Sun21:Markov, Park2021919, Savas21, Park2018387, Lee22:Optimal, Lee22:Urban, Kim2019, Ahmed20171792}, long-term evolution (LTE) signals \cite{Shamaei21, Maaref20, Kang20191182, Jeong2020958, Lee2020:Preliminary, Lee2020939, Lee20202347, Jia21:Ground, Kang2020774, Lee22:Evaluation}, enhanced long-range navigation (eLoran) system \cite{Kim22:First, Son20191828, Son2018666, Williams13, Kim2020796, Qiu10, Park2020824, Li20, Rhee21:Enhanced}, and various other techniques \cite{Kim2017:SFOL, Lee22:SFOL, Rhee2019, Park2020800, Kang21:Indoor}. Additionally, depth cameras, RGB cameras, lidar, and radar can be installed on UAVs for collision avoidance. UAVs can be used in search missions, aerial photography, delivery, airspace management, and communications relays \cite{Sibanyoni19, Dorling17, Hiraguri20, Ho19}.

Owing to the wide use-cases of UAVs, controlling them autonomously using artificial intelligence (AI) techniques has been studied extensively \cite{Kouroshnezhad21, Chhikara21, Kim2020784, Lai21}. Among them, reinforcement learning is used to learn the optimal behavior in each situation through a reward. It is mainly used in the fields of robotics and game AI and has been widely used owing to recent developments in deep learning.

In this study, we used reinforcement learning on two quadcopters to find the fastest route to a target location while avoiding obstacles. To reduce learning time and cost, and eliminate the risk of damage to the real quadcopter, learning was conducted through simulation in a virtual environment. The indoor virtual environment was implemented using Gazebo \cite{Koenig04}, which is an open-source 3D robotics simulator. Additionally, to improve learning performance, curriculum learning \cite{Bengio09} was conducted in two stages.

A curriculum learning method teaches an easier task first before teaching a more difficult and complex activity. This approach has benefit in terms of generalization and convergence time. In the first stage of our study, a simple path to fly to a target location in the shortest time in an obstacle-free environment was learned. Thereafter, in the second stage, a relatively complex task of flying to the target location in an obstacle-added environment was learned. Curriculum learning enables more efficient learning compared to learning complex tasks from the beginning.

When other agents are considered as part of the multi-agent environment, there are non-stationary characteristics of the multi-agent environment wherein the optimal behavior varies based on the action of other agents \cite{Jang19}. Thus, we added the action of the other agent to the state space, which is a common practice to consider the non-stationary characteristics \cite{Georgios19}.

\section{Virtual simulation}

Our simulation of multiple UAVs in an indoor virtual environment was implemented using Robot Operating System (ROS) \cite{Quigley09}, Gazebo \cite{Koenig04}, OpenAI Gym \cite{Brockman16}, and RLlib.
ROS is an open source meta-operating system for robotics applications \cite{Quigley09}. It was used in this study because it is widely used in the robotics field and can be easily combined with various robot software frameworks. 
By using the topic communication, which is a message-passing method within ROS, sensor and location information obtained by each UAV were introduced into the learning environment of itself or of the other UAV.

\begin{figure}
  \centering
  \centerline{\includegraphics[width=0.9\linewidth]{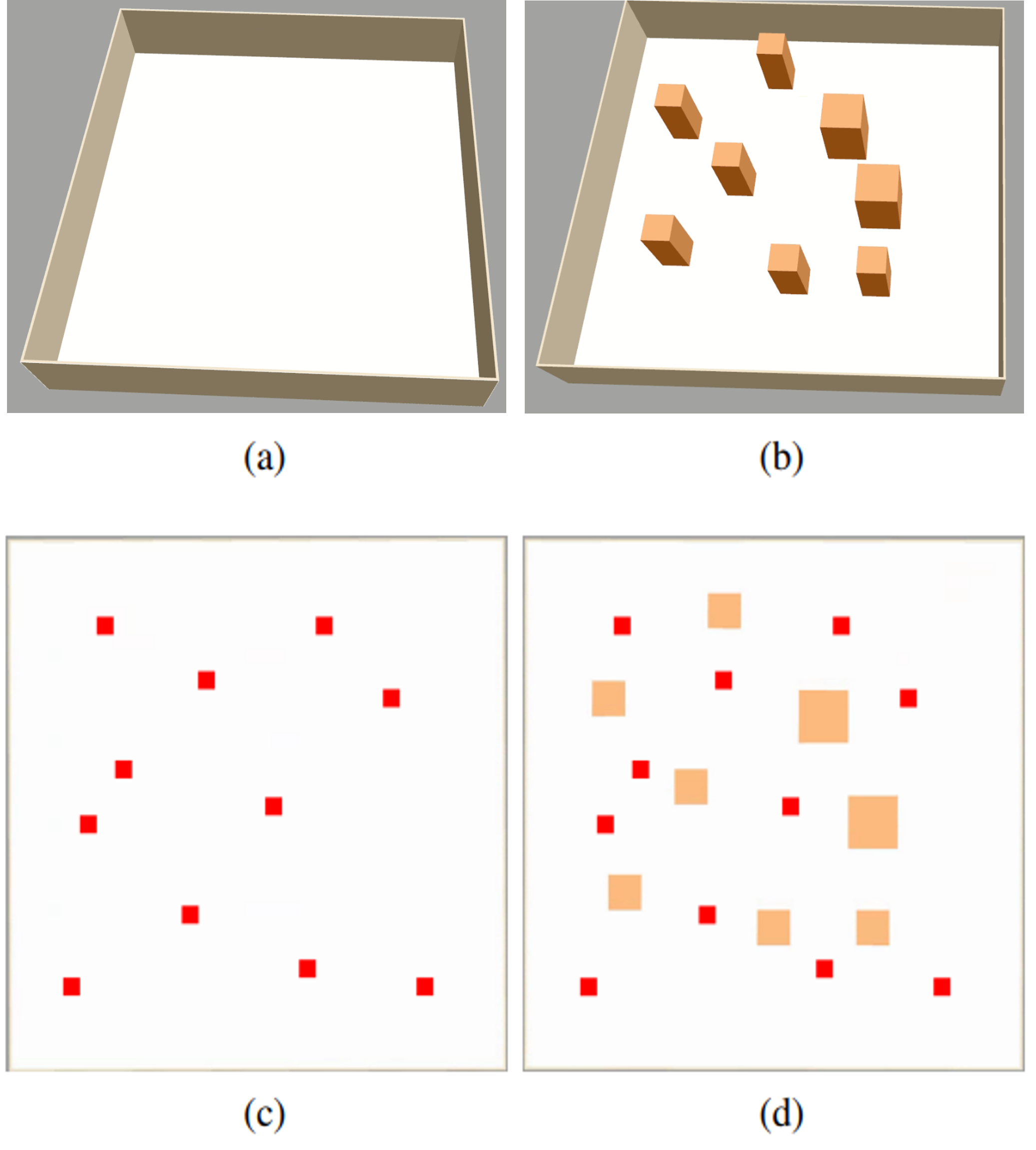}}
  \caption{Indoor virtual environment}
  \label{fig:indEnv}
\end{figure}

To implement an indoor virtual environment, a single-floor structure of 30 m $\times$ 30 m was created using Gazebo's internal tool, the building editor. 
Figs. \ref{fig:indEnv}(a) and \ref{fig:indEnv}(b) show the obstacle-free environment used in the first curriculum learning stage and the obstacle-added environment used in the second curriculum learning stage, respectively. 
Figs. \ref{fig:indEnv}(c) and \ref{fig:indEnv}(d) illustrate the start and goal position candidates for the UAVs, which are indicated by red squares.

OpenAI Gym is an open-source Python library that contains reinforcement learning  algorithms \cite{Brockman16}. 
In OpenAI Gym, various standard learning environments are provided as references.
A new learning environment was implemented in this study according to the learning environment format of OpenAI Gym.
OpenAI Gym can be easily linked with PyTorch, TensorFlow, or RLlib \cite{Liang18} which is a reinforcement learning algorithm library.

\section{Learning environment}

The learning time was divided into \textit{step}, \textit{episode}, and \textit{iteration}. 
A step is defined as the time to choose an action and obtain a reward and it is the minimum time unit. 
An episode refers to the time required for the task to succeed or fail, and 
an iteration updates the model parameters when the episodes have sufficiently progressed.

The size of one iteration is determined by the \textit{train batch size} parameter, which was selected as 20,000 steps in this study.
The first stage of the curriculum learning was progressed for 150 iterations, and thereafter the second stage was continuously learned for 310 iterations.

Proximal policy optimization (PPO) \cite{Schulman17}, which is a reinforcement learning algorithm based on the policy-gradient method, was used in this study because it is more suitable than reinforcement learning algorithms based on Q-learning such as deep Q-networks \cite{Volodymyr13} for situations with a continuous state space \cite{Jang19, Klissarov17}. 
The RLlib provides the PPO algorithm under the name of PPOTrainer. 
We used the RLlib and hyper-parameter settings of \cite{Park21:Indoor}.

\subsection{State space}

The state space was divided into two parts and the heading, distance, and lidar data were used to determine the environment. The heading to the other UAV, distance to the other UAV, relative heading of the other UAV, and current action of the other UAV were used to determine its situation. 
In this study, \textit{heading} means the counterclockwise rotation angle required for the UAV to look at the goal position (in radians). 
\textit{Distance} means the distance that the UAV needs to move when it is looking at the goal position (in meters).

Considering the non-stationary characteristics in a multi-agent environment, the information and actions of the other UAV were provided, and to determine the differences in learning performance, two models with different state spaces were trained and compared.
One included only heading, distance, and lidar data of each UAV in the state space of the corresponding UAV, and the other included additional information and actions of the other UAV in the state space of each UAV.

\subsection{Action space}

The action space consisted of forward speeds and yaw rates. 
The basic forward speed and yaw rates were set to 0.5 m/s, and $\frac{\pi}{12}$ rad/s, respectively. The three forward speeds were 0, 1, and 2 times the default forward speed, and the five yaw rates were $-2$, $-1$, 0, 1, and 2 times the default yaw rate. 
A negative yaw rate means counterclockwise rotation, and a positive yaw rate means clockwise rotation. 
Because this study assumed a 2D situation, the vertical velocity was set to 0.

\subsection{Reward model}

Among the two reward models used in a previous study \cite{Park21:Indoor}, the one that showed good learning performance was applied to the current study.
At the end of each episode, a large positive reward is given if the task succeeds; a large negative reward is given if the task fails.
To implement movement towards the goal position as quickly as possible, a negative reward is given to each step.
In addition, there are rewards regarding the distance and heading to the goal position. 
The detailed reward model is given in \cite{Park21:Indoor}.

\section{Simulation results}

In this study, three learning models were trained. Model 1 did not proceed with curriculum learning but learned directly in the obstacle-added environment. Model 2 performed curriculum learning but did not include information and actions of the other agent in the state space. Model 3 included information and actions of the other agent in the state space, and performed curriculum learning.

\begin{figure}
  \centering
  \centerline{\includegraphics[width=0.9\linewidth]{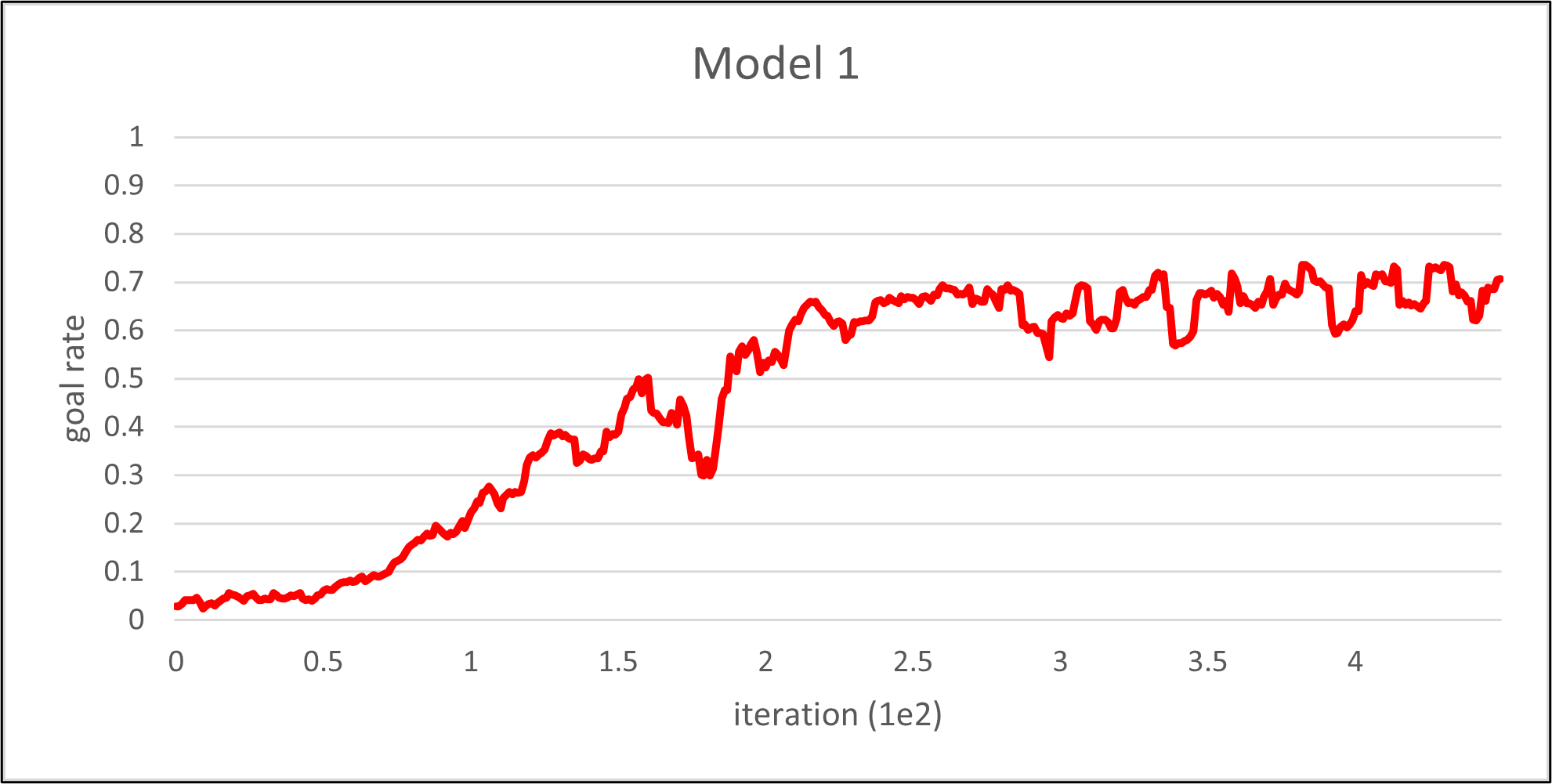}}
  \caption{Goal rate for Model 1 as a moving average value}
  \label{fig:Model1}
\end{figure}

Fig. \ref{fig:Model1} shows the goal rate of each iteration for Model 1 as a moving average value. Model 1 was trained for 460 iterations in an environment with obstacles. The goal rates of the previous ten iterations were used to calculate a moving average value. A maximum goal rate of 73.6\% was attained by Model 1.

\begin{figure}
  \centering
  \centerline{\includegraphics[width=0.9\linewidth]{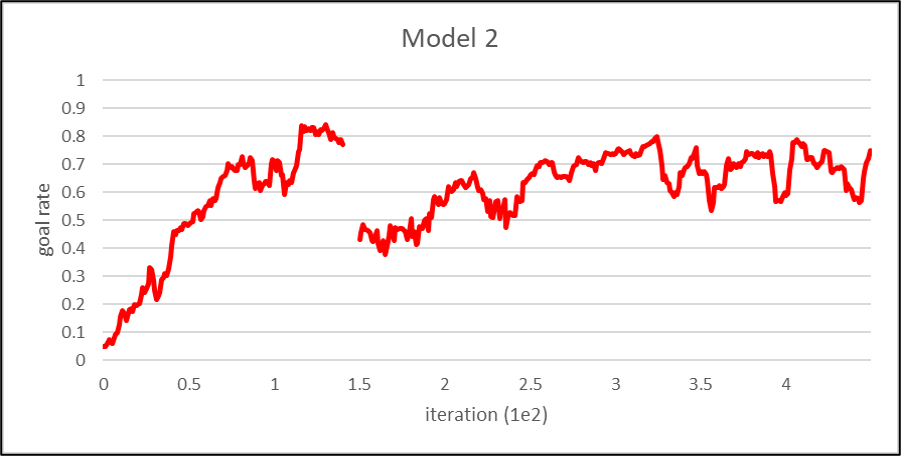}}
  \caption{Goal rate for Model 2 as a moving average value}
  \label{fig:Model2}
\end{figure}

\begin{figure}
  \centering
  \centerline{\includegraphics[width=0.9\linewidth]{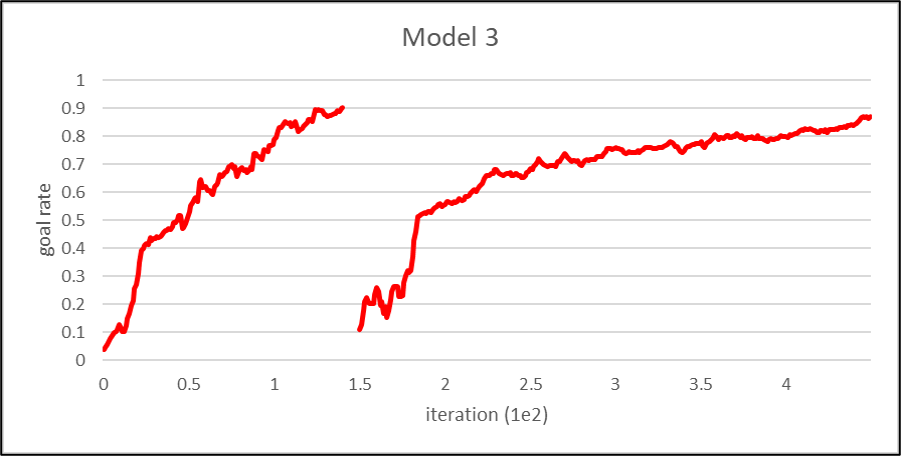}}
  \caption{Goal rate for Model 3 as a moving average value}
  \label{fig:Model3}
\end{figure}

Figs. \ref{fig:Model2} and \ref{fig:Model3} show the goal rate of each iteration for Models 2 and 3, respectively, as a moving average value. Models 2 and 3 were trained for 150 iterations in the first curriculum learning stage and additionally trained for 310 iterations in the second stage.
Maximum goal rates of 79.9\% and 87.0\% were attained by Models 2 and 3, respectively, in an environment with obstacles.

Model 1, which did not use curriculum learning, directly learned in the obstacle-added environment, and achieved the lowest goal rate. In contrast, Models 2 and 3, which used curriculum learning, achieved relatively high goal rates. 
Model 2 showed unstable learning progress, which is indicated by the fluctuations of the goal rates in Fig. \ref{fig:Model2}, because of the non-stationary characteristics in a multi-agent environment. 
Whereas, Model 3, which added information and actions of the other UAV to the state space to consider the non-stationary characteristics, showed stable learning progress compared with Model 2. 
The maximum goal rates achieved by Models 1, 2, and 3 were 73.6\%, 79.9\%, and 87.0\%, respectively. Therefore, Model 3 performed significantly better than the other two.

\section{Conclusion}

In this study, we performed reinforcement learning for multi-UAV indoor path planning. Curriculum learning was performed in two stages to increase learning efficiency. 
We showed that the cases with the curriculum learning demonstrated higher goal rates than the other case after the same number of iterations.
It was also shown that it is important to include the other agent's information in the state space when a multi-agent environment is considered. 
When the other agent's information was included in the state space, more stable learning progress and higher goal rates were achievable.

\section*{Acknowledgment}

This research was supported by the Unmanned Vehicles Core Technology Research and Development Program through the National Research Foundation of Korea (NRF) and the Unmanned Vehicle Advanced Research Center (UVARC) funded by the Ministry of Science and ICT, Republic of Korea (2020M3C1C1A01086407).
This work was also supported by the Institute of Information \& Communications Technology Planning \& Evaluation (IITP) grant funded by the Korea government (KNPA) (2019-0-01291).

\bibliographystyle{IEEEtran}
\bibliography{output.bbl}

\vspace{12pt}

\end{document}